\pdfoutput=1
\documentclass[letterpaper, 10 pt, conference]{ieeeconf}  % Comment this line out if you need a4paper

\IEEEoverridecommandlockouts                              % This command is only needed if 
\usepackage{amsmath} % assumes amsmath package installed
\usepackage{amssymb}  % assumes amsmath package installed

\usepackage{bm}
\usepackage{cite}
\usepackage{flushend}
\include{preamble}

\title{\LARGE \bf
    Vision-Language Interpreter for Robot Task Planning 
}

\author{
    Keisuke Shirai$^{1*}$, 
    Cristian C. Beltran-Hernandez$^{2}$,
    Masashi Hamaya$^{2}$, 
    Atsushi Hashimoto$^{2}$, \\
    Shohei Tanaka$^{2}$,
    Kento Kawaharazuka$^{3}$, 
    Kazutoshi Tanaka$^{2}$,
    Yoshitaka Ushiku$^{2}$,
    Shinsuke Mori$^{4}$% <- this % stops a space
    \thanks{$^{1}$Keisuke Shirai and Shinsuke Mori are with Kyoto University, Kyoto 606-8501, Japan (email: [shirai.keisuke.64x@st.kyoto-u.ac.jp,forest@i.kyoto-u.ac.jp])}% <- this % stops a space
    \thanks{$^{2}$Cristian C. Beltran-Hernandez, Masashi Hamaya, Atsushi Hashimoto, Shohei Tanaka, Kazutoshi Tanaka, Yoshitaka Ushiku are with the OMRON SINIC X Corporation, Tokyo 113-0033, Japan (email: [cristian.beltran, 
    masashi.hamaya, atsushi.hashimoto, shohei.tanaka, kazutoshi.tanaka, yoshitaka.ushiku]@sinicx.com)}% <- this % stops a space
    \thanks{$^{3}$Kento Kawaharazuka is with the University of Tokyo, 73-1 Hongo, Bunkyo-ku, Tokyo, 113-8656, Japan (email: kawaharazuka@jsk.t.u-tokyo.ac.jp)}% <- this % stops a space
    \thanks{$^*$Work was done while the first author was an intern at OMRON SINIC X Corporation.}% <- this % stops a space
}

\date{\vspace{-2em}}

\begin{document}

\maketitle
\thispagestyle{empty}
\pagestyle{empty}

%%%%% Abstract
\begin{abstract}
Large language models (LLMs) are accelerating the development of language-guided robot planners. Meanwhile, symbolic planners offer the advantage of interpretability. This paper proposes a new task that bridges these two trends, namely, \textit{multimodal planning problem specification}. The aim is to generate a problem description (PD), a machine-readable file used by the planners to find a plan. By generating PDs from language instruction and scene observation, we can drive symbolic planners in a language-guided framework. We propose a Vision-Language Interpreter (ViLaIn), a new framework that generates PDs using state-of-the-art LLM and vision-language models. ViLaIn can refine generated PDs via error message feedback from the symbolic planner. Our aim is to answer the question: How accurately can ViLaIn and the symbolic planner generate valid robot plans? To evaluate ViLaIn, we introduce a novel dataset called the problem description generation (ProDG) dataset. The framework is evaluated with four new evaluation metrics. Experimental results show that ViLaIn can generate syntactically correct problems with more than 99\% accuracy and valid plans with more than 58\% accuracy. Our code and dataset are available at \url{https://github.com/omron-sinicx/ViLaIn}.
\end{abstract}

%%%%% Introduction 
\section{INTRODUCTION}\label{sec:introduction}
Natural language is a prospective interface for non-experts to instruct robots intuitively~\cite{hatori2018interactively,tellex2020robots,liang2023code}. Earlier studies have used recurrent neural networks~\cite{hochreiter1997long,cho2014learning} to map abstract linguistic instructions to representations for robots~\cite{arumugam2017accurately,hatori2018interactively,paxton2019prospection}. Here, the linguistic instructions represent desired goal conditions. More recent studies use large language models (LLMs)~\cite{touvron2023llama,openai2023gpt,anil2023palm} to directly generate robot plans from the instructions~\cite{huang2022language,raman2022planning,lin2023text2motion,singh2023progprompt}. These language-guided planners utilize few-shot prompting to solve tasks without training~\cite{brown2020language}. The plans are a sequence of discrete symbolic actions (e.g., \texttt{pick(a)} and \texttt{place(a, b)}) that complete the task. We aim to strengthen the language-guided planners in terms of the improvement of interpretability.\footnote{We define interpretability as a mechanism to provide insights into the inner workings of the system.} Interpretability is essential to gain the trust of the user and provide insights into the robot's decision-making process~\cite{gilpin2018explaining}. For example, the identification of failure causes through interpretation leads to continuous improvement of overall performance. 

\begin{figure}[t]
    \centering
    \includegraphics[width=0.9\columnwidth]{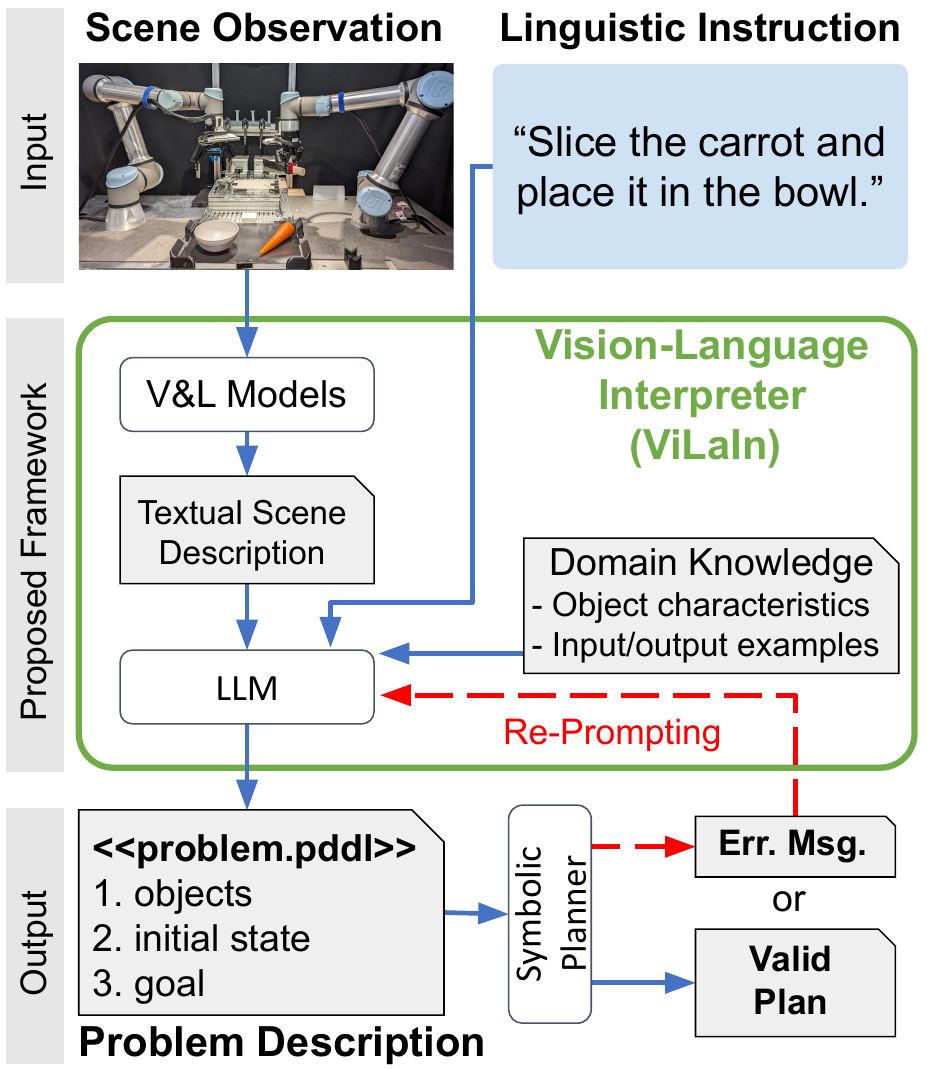}
    \caption{Overview of our approach. The vision-language interpreter (ViLaIn) generates a problem description from a linguistic instruction and scene observation. The symbolic planner finds an optimal plan from the generated problem description.}
    \label{fig:overview}
\end{figure}

Robot task planning has traditionally been solved using symbolic planning~\cite{karpas2020automated}. Modern symbolic planners use the Planning Domain Definition Language (PDDL) to describe planning problems. In PDDL, a planning problem is defined in two parts: the \textit{domain} that defines the state of variables and actions, and a \textit{problem description} (PD) that defines the objects of interest, their initial state, and the desired goal state~\cite{haslum2019pddl,fox2003pddl2}. The domain and problem are inputs to the planner to find an optimal plan, a sequence of symbolic actions. 

Symbolic planners offer several benefits. The domain and problem descriptions are human-readable, especially when variable names are chosen intuitively. Moreover, the obtained plans are guaranteed to be logically correct. Considering these advantages, combining symbolic planning and language-guided planning is a promising research direction to realize interpretable robots. To that end, we proposed generating the PDs from natural language instructions. 
Since the linguistic instructions only represent the goal conditions, additional information about the environment is required to generate the initial state (e.g., an image representing the current environment). We refer to this additional information as \textit{scene observations}.

We tackle the \textit{multimodal planning problem specification} task, a new task for transforming linguistic instructions and scene observations into logically and semantically correct PDs. The PDs have to be executable by the symbolic planners. This paper investigates how accurately we can generate such PDs with a state-of-the-art LLM~\cite{openai2023gpt} and vision-language model~\cite{liu2023grounding,li2023blip2} without additional training. We propose a Vision-Language Interpreter (ViLaIn), a new framework to solve the PD generation task, illustrated in \figref{fig:overview}. ViLaIn consists of three modules that generate each part of the PDs. The complete PD is assembled by concatenating these parts. Furthermore, ViLaIn can refine the generated PDs via error feedback from the symbolic planner. The planner uses a pair of the generated PD and the domain description to find a plan. We use the state-of-the-art symbolic planner called Fast Downward~\cite{helmert2006fast} throughout this paper. 

\begin{table}[t]
    \centering
    \caption{Differences between previous studies and ours} 
    \begin{tabular}{l|c|l}\hline
        \multicolumn{1}{c|}{\multirow{2}{*}{Approach}} & Input other than & \multicolumn{1}{c}{\multirow{2}{*}{Output}} \\
        & linguistic instruction & \\\hline
        Huang et al.~\cite{huang2022language} & --- & Symbolic action \\
        Raman et al.~\cite{raman2022planning} & --- & Symbolic action \\
        Text2Motion~\cite{lin2023text2motion} & PDDL scene desc. & Symbolic action \\
        SayCan~\cite{ahn2022can} & Image & Pre-defined skill \\
        RT-2~\cite{brohan2023rt} & Image & Low-level action \\
        ProgPrompt~\cite{singh2023progprompt} & --- & Program code \\
        Code as Policies~\cite{liang2023code} & Image & Program code \\
        LLM+P~\cite{liu2023llm} & Linguistic scene desc. & Problem desc. \\
        ViLaIn (ours) & Image & Problem desc. \\\hline
    \end{tabular}
    \label{tab:difference}
\end{table}

To evaluate ViLaIn, we introduce a novel dataset called the problem description generation (ProDG) dataset. The ProDG dataset consists of linguistic instructions, scene observations, and domain and problem descriptions. The descriptions are written in PDDL~\cite{fox2003pddl2}. This dataset covers three domains: cooking as a practical robot domain, and the blocks world and the tower of Hanoi as classical planning domains. We propose four new evaluation metrics to evaluate ViLaIn from multiple perspectives.

The main contributions of this work are three-fold:
\begin{itemize}
    \item Multimodal planning problem specification, a new task to bridge the language-guided planning and symbolic planners with scene observations. 
    \item Vision-Language Interpreter (ViLaIn), a new framework consisting of a state-of-the-art LLM and vision-language model. ViLaIn can refine erroneous PDs by using error messages from the symbolic planner.
    \item The problem description generation (ProDG) dataset, a new dataset that covers three domains: the cooking domain, the blocks world, and the tower of Hanoi. The dataset comes with new metrics that evaluate ViLaIn from multiple perspectives.
\end{itemize}

%%%%% Related Work
\section{RELATED WORK}\label{sec:related}
This section describes previous work on language-guided planning, symbolic planning, and scene recognition in computer vision. \tabref{tab:difference} summarizes the difference between several studies mentioned here and ViLaIn. 

\subsection{Planning from Natural Language}
Task planning from natural language has been actively studied~\cite{huang2022language,ahn2022can,liu2023grounding}. Converting linguistic instructions into symbolic actions via neural networks is a typical approach~\cite{paxton2019prospection,sharma2022skill}. More recent studies~\cite{huang2022language,raman2022planning,lin2023text2motion,singh2023progprompt} use LLMs and directly generate plans with few-shot prompting~\cite{brown2020language}. However, these language-guided planners have two issues. First, their systems hide the inner workings by generating plans end-to-end. Second, the obtained plans are not guaranteed to be logically correct. ViLaIn resolves these issues by converting instructions into human-readable PDs and driving symbolic planners to find plans with the generated PDs. A recent study uses LLMs to convert linguistic instructions and images into programs to complete robot tasks~\cite{liang2023code}. PDs describe tasks more specifically, and their logical correctness is automatically verifiable. In other words, ViLaIn has the potential to deliver validated machine-readable information to other language-guided planners as an auxiliary input. 

More recent studies have used LLMs to convert natural language inputs to PDs~\cite{liu2023llm,xie2023translating}. However, one study~\cite{liu2023llm} assumes that scene descriptions (the objects and initial state) are provided in natural language, which is not practical for real applications. Another work~\cite{xie2023translating} focuses on only generating the goal specifications. Contrary to these studies, ViLaIn uses images for scene descriptions and generates the whole PDs, including the objects and initial states.

\subsection{Symbolic Planning with PDDL}
Symbolic planning (automated planning) has been used to solve robotic tasks~\cite{karpas2020automated}. Symbolic planners~\cite{bonet2001planning,helmert2006fast} use domain and problem descriptions to find plans, which are sequences of (symbolic) actions that alter the environment from its initial state to a goal state. The descriptions are written in formal languages, such as PDDL~\cite{fox2003pddl2} and PDDLStream~\cite{garrett2020pddlstream}. Robots execute low-level actions based on the found high-level plans of PDDL~\cite{ahmadzadeh2015learning,wang2021learning,silver2021learning}. This framework enables robots to solve various problems but assumes a preparation of corresponding PD for each problem. ViLaIn is designed to collaborate with those PDDL-based planning frameworks by translating linguistic instructions into PDs.

\subsection{Scene Recognition for Planning Problem Specification}
The generation of the objects and initial state in PD is related to research in computer vision. This section briefly overviews such previous work.

The object part of PDs lists objects required for the task. This work generates the objects from scene observations. This can be viewed as object detection in computer vision. Classical object detectors~\cite{ren2015faster,redmon2016you} have been developed focusing on a fixed number of classes (e.g., person and dog). However, our task handles objects not included in the classes. Hence, we use an open-vocabulary object detector~\cite{zareian2021open,liu2023grounding}. These detectors have recently gained attention because they can detect arbitrary objects using text queries.

The \textit{initial state} represents object relationships and their states. Detecting such scene descriptions from images has been addressed on visual relationship detection~\cite{lu2016visual,inayoshi2020bounding} or scene graph generation~\cite{xu2017scene,yang2022panoptic}. Previous work trained a model with PDDL predicates and demonstrated it in real robot domains~\cite{migimatsu2022grounding}. We use a state-of-the-art LLM and vision-language model to generate the initial state.

%%%%% Problem Statement
\section{PROBLEM STATEMENT}\label{sec:task}
We focus on multimodal planning problem specification, a new task for bridging language-guided planning and symbolic planning. The input is a quadruple $(L, S, D_D, D_K)$; a linguistic instruction $L$, a scene observation $S$, a domain description $D_D$, and domain knowledge $D_K$. $L$ is a sequence of words describing the task. $S$ is an RGB image describing the initial state of the environment. $D_D$ defines parts common to all problems: object types (e.g., \texttt{location} and \texttt{tool}), predicates (e.g., \texttt{at} and \texttt{clear}), and symbolic actions (e.g., \texttt{slice} and \texttt{pick}). $D_K$ supports $D_D$ by providing more specific information on each problem, such as object characteristics (e.g., the cutting board is round, the counter is black) and actual input/output examples. Note that the examples in $D_K$ use the object types and predicates defined in $D_D$.

The output is a PD $P$ consisting of $(O, I, G)$: the objects $O$, the initial state $I$, and the goal specification $G$. $O$ consists of objects required for the task completion (e.g., \textit{carrot} and \textit{knife}). $I$ consists of a set of propositions that represent the initial state of the environment (e.g., \texttt{(at carrot counter)}). A proposition is formed by providing a predicate with arguments. For example, providing a predicate \texttt{(at ?a1 ?a2)} with (\texttt{a1}, \texttt{a2}) = (\texttt{carrot}, \texttt{cutting\textunderscore board}) forms a proposition \texttt{(at carrot cutting\textunderscore board)} meaning "the carrot is at the cutting board." $G$ consists of a set of propositions that represent the desired goal condition of the environment. For example, \texttt{(and (at carrot bowl) (is-sliced carrot))} represents the goal condition that "the carrot should be sliced and should be at the bowl." $P$ and $D_D$ are written in PDDL~\cite{fox2003pddl2}, following previous work~\cite{liu2023llm,xie2023translating}. We refer to $O$, $I$, or $G$ with PDDL (e.g., the PDDL objects). The goal of this task is obtaining a function $M: (L, S, D_D, D_K) \rightarrow (O, I, G)$. $P$ must be machine-readable and executable by the symbolic planner.

%%%%% Vision-Language-Interpreter (ViLaIn)
\begin{figure}[t]
    \centering
    \includegraphics[width=0.85\columnwidth]{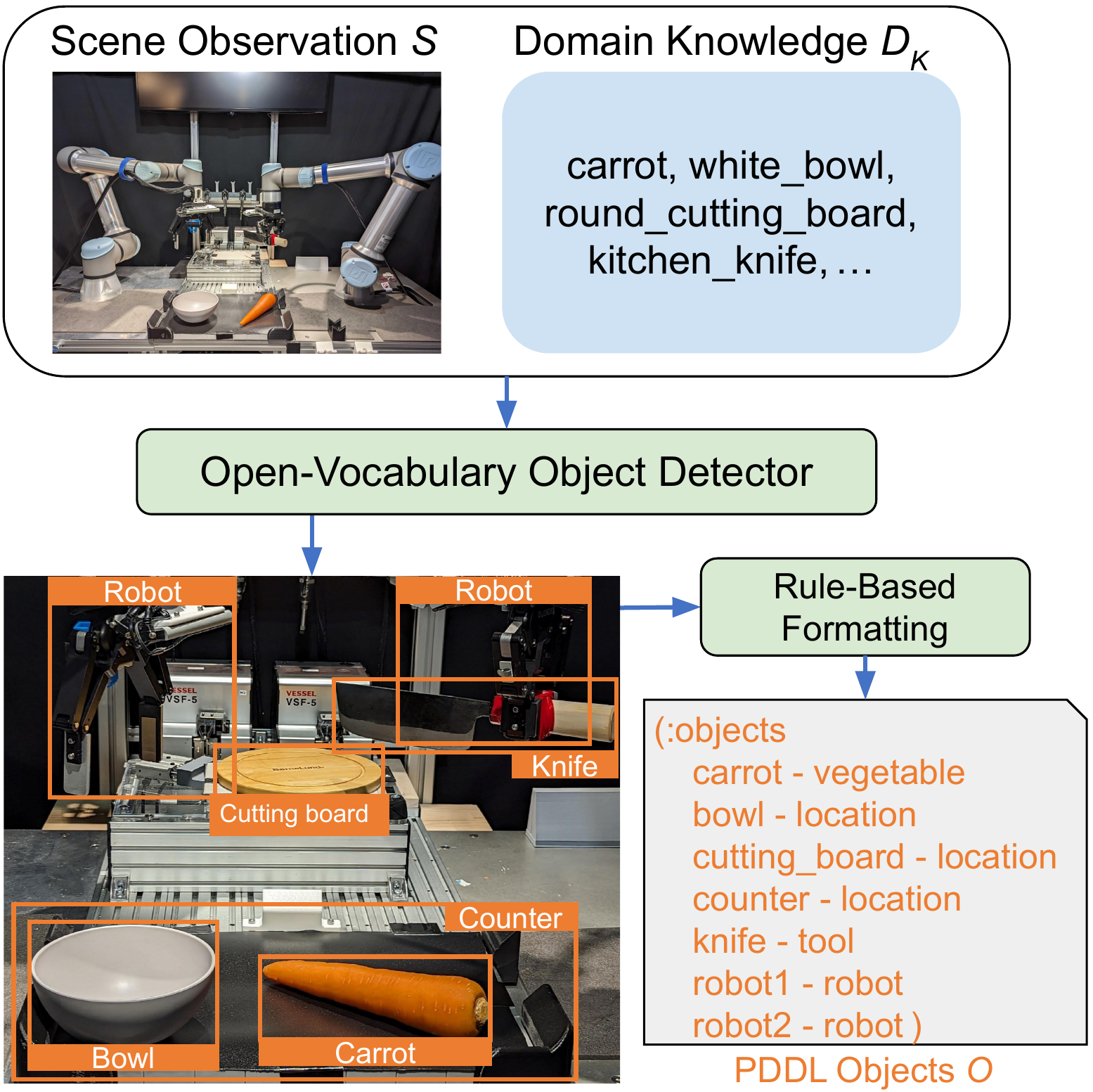}
    \caption{The open-vocabulary object detector detects objects from the observation. The text query is provided by the domain knowledge. The detected objects are converted into a PDDL format in a rule-based way.}
    \label{fig:object-estimator}
\end{figure}

\section{Vision-Language Interpreter}\label{sec:approach}
ViLaIn consists of three modules: the object estimator, the initial state estimator, and the goal estimator. We describe these modules in this section.

\subsection{Object Estimator}\label{subsec:object-estimator}
The PDDL objects $O$ list objects of interest in the scene observations $S$. However, the observed objects vary greatly from domain to domain. Further, it must recognize various objects that classical object detectors cannot handle. For this reason, we use Grounding-DINO~\cite{liu2023grounding}, a state-of-the-art open-vocabulary object detector. \figref{fig:object-estimator} illustrates the estimator. We assume that the list of objects for the task is known. The object list can be used as the text query. However, we found from preliminary experiments that simply using the object list fails to detect several objects. To address this issue, we elaborate the query using the domain knowledge (e.g., "cutting board" $\rightarrow$ "round cutting board" and "knife" $\rightarrow$ "kitchen knife").\footnote{In this work, we assume that the domain knowledge is created by humans, and we leave the automatic generation of it to future work.} In our setting, these elaborated queries are included in the domain knowledge $D_K$. The detected objects are converted into a PDDL format by rules. 

\begin{figure}[t]
    \centering
    \includegraphics[width=0.8\columnwidth]{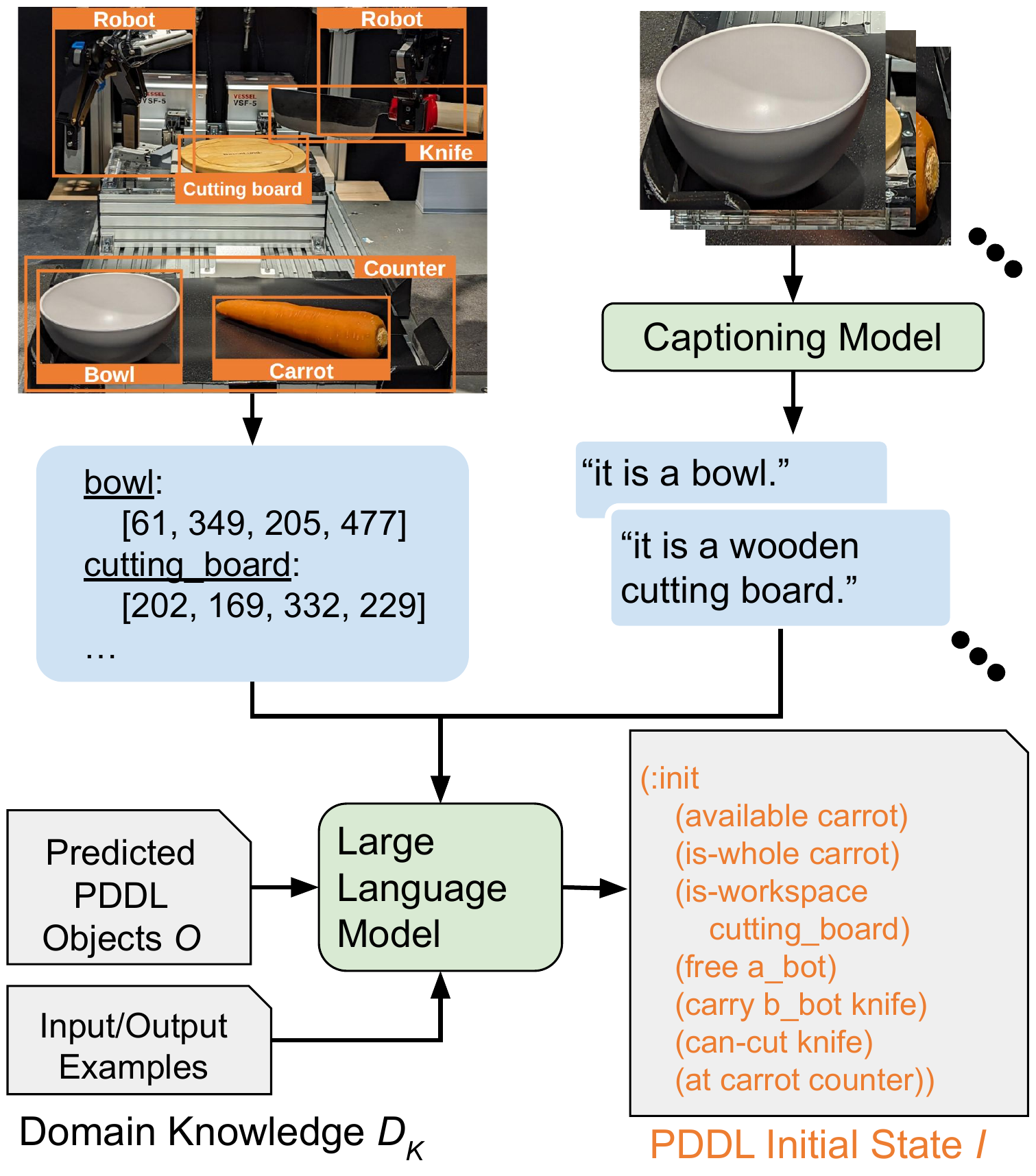}
    \caption{The captioning model generates captions for each object. The LLM generates the PDDL initial state from the bounding boxes and the captions using few-shot prompting.}
    \label{fig:initial-state-estimator}
\end{figure}

\subsection{Initial State Estimator}\label{subsec:initial-state-estimator}
The PDDL initial states $I$ must specify the initial state of the environment using propositions. Here, different predicates from $D_D$ should be used for different domains to represent the propositions. In addition, omitting a single proposition could cause an invalid PD by making reaching the goal from the initial state impossible. We implement the initial state estimator with a combination of an LLM and image captioning model. \figref{fig:initial-state-estimator} shows the estimator. We use BLIP-2~\cite{li2023blip2} as the captioning model and GPT-4~\cite{openai2023gpt} as the LLM. Given the objects' bounding boxes, BLIP-2 generates captions for each object with a prompt of "Q: what does this {object} describe? A: ." GPT-4 generates the PDDL initial state $I$ from the bounding boxes and captions. GPT-4 uses few-shot prompting and leverages input/output examples in $D_D$ to derive available predicates.

\subsection{Goal Estimator}\label{subsec:goal-estimator}
\begin{figure}[t]
    \centering
    \includegraphics[width=0.85\columnwidth]{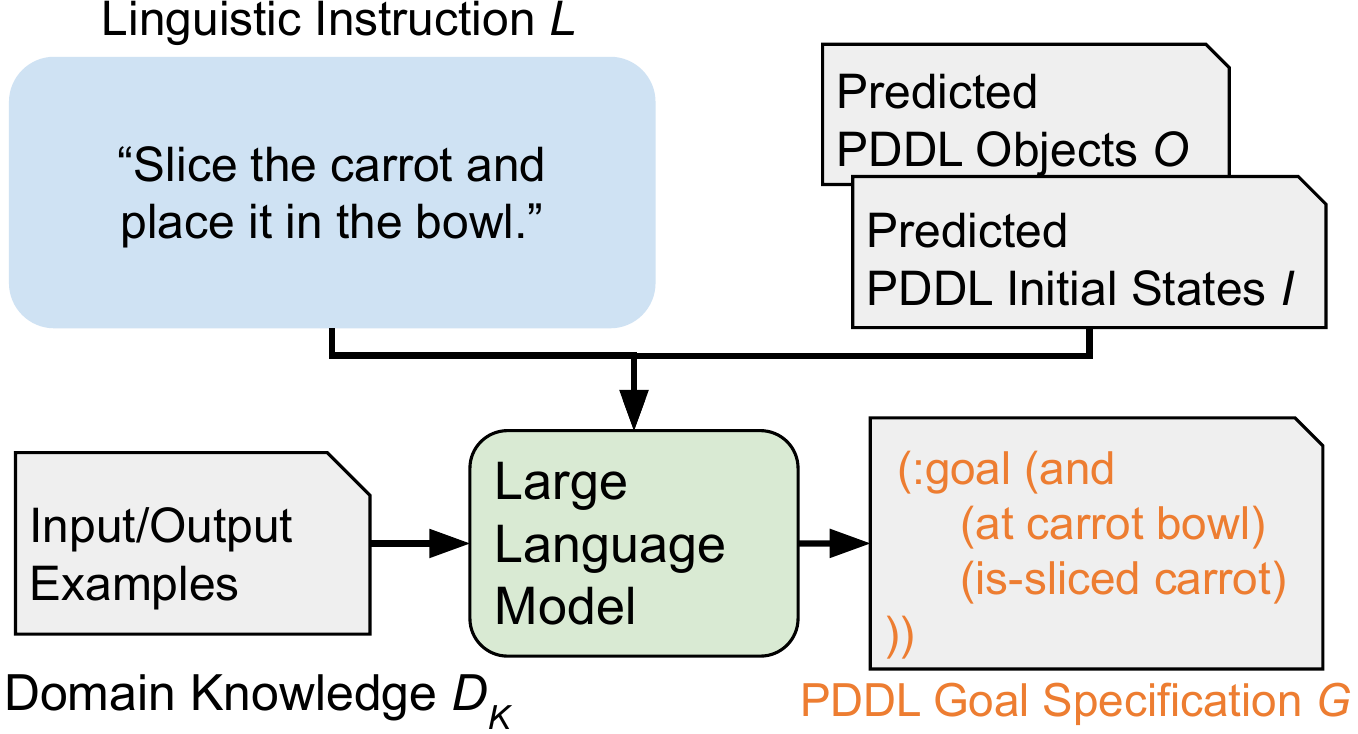}
    \caption{The LLM directly generates the PDDL goal specification from the instruction and the PDDL objects and initial state using few-shot prompting.}
    \label{fig:goal-estimator}
\end{figure}

The PDDL goal specifications $G$ must represent the desired goal conditions specified by the linguistic instructions $L$. Generating $G$ requires $O$ to refer to the object list and $I$ to consider the relationships of the objects. We implement the goal estimator with an LLM, following previous work~\cite{lin2023text2motion,xie2023translating}. \figref{fig:goal-estimator} shows the estimator. We use GPT-4 to generate $G$ from $L$, $O$, and $I$. Similarly to \secref{subsec:initial-state-estimator}, GPT-4 uses few-shot prompting with $D_K$. 

\begin{table*}[t]
    \centering
    \caption{Defined object types, predicates, and actions in the domain descriptions}
    \begin{tabular}{l|c|c|c}\hline
        Domain & Object types & Predicates & Actions \\\hline
        \multirow{2}{*}{Cooking} & vegetable, location, & available, is-whole, is-sliced, free, & \multirow{2}{*}{pick, place, slice} \\
        & tool, robot & carry, can-cut, at, at-workspace & \\\hline
        Blocksworld & block, robot & on, ontable, clear, handempty, handfull, holding & pick-up, put-down, stack, unstack \\\hline
        Hanoi & disk, peg & clear, on, smaller, move & move \\\hline
    \end{tabular}
    \label{tab:domain}
\end{table*}

\subsection{Corrective Re-Prompting}\label{subesc:corrective-reprompting}
Generated PDs are used by the planner to find plans. The planning might fail in the following two cases. One is when the PDs are syntactically incorrect. Generating propositions with undefined objects in $O$ or undefined predicates in $D_D$ results in such PDs (e.g., create \texttt{(at cucumber counter)}, but \texttt{cucumber} is not listed in $O$). The other is when the generated $O$ is unreachable from the generated $I$. Contradictory propositions create such a PD (e.g., both of a proposition \texttt{(on red\textunderscore block blue\textunderscore block)} and the opposite one \texttt{(on blue\textunderscore block red\textunderscore block)} exist in $I$). In both cases, the planner stops planning and returns an error message, a clue to refine the erroneous parts. It is ideal if the system automatically refines the PDs via the error messages. ViLaIn has such a mechanism, and we describe it in this section.

\begin{figure}[t]
    \centering
    \includegraphics[width=0.95\columnwidth]{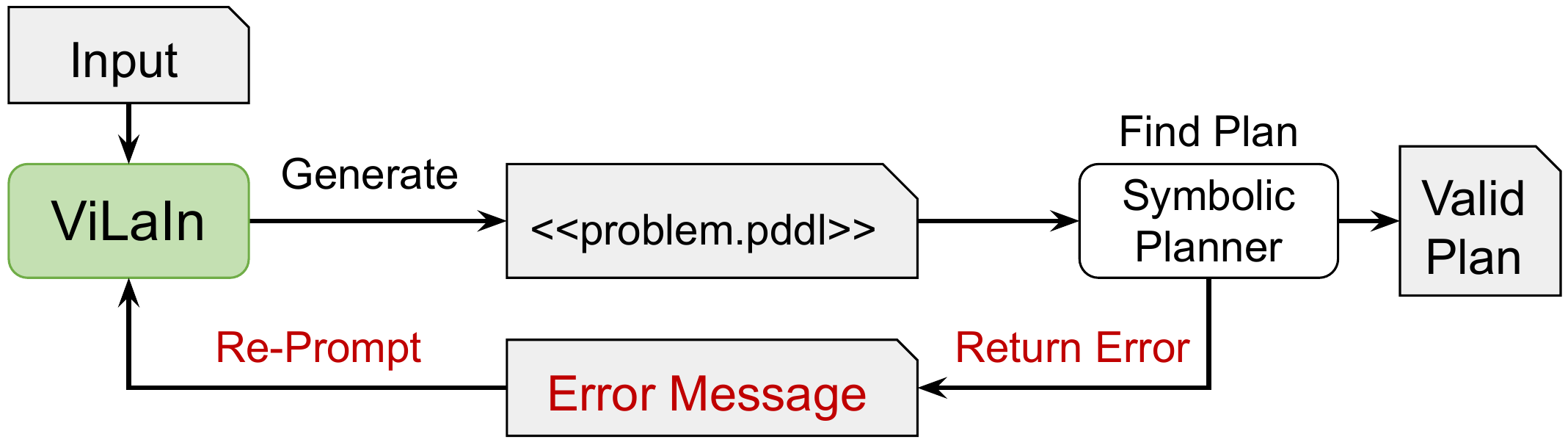}
    \caption{ViLaIn can refine the generated problem description via an error message from the planner.}
    \label{fig:corrective-reprompting}
\end{figure}

When the planning fails, ViLaIn creates a prompt and re-prompts GPT-4 to refine the PD. We refer to this technique as Corrective Re-prompting (CR), following previous work~\cite{raman2022planning}. \figref{fig:corrective-reprompting} shows ViLaIn with CR. The prompt consists of input/output examples in $D_K$, the current input ($L$ and $S$), the generated problem $P$, and the error message. 

\paragraph*{\textbf{Chain-of-Thought prompting}}
We use Chain-of-thought (CoT) prompting~\cite{wei2022chain,kojima2022large,zhou2023least} to further strengthen CR. CoT is a technique for solving complex reasoning tasks by LLMs. CoT introduces an intermediate reasoning step before generating the final output. With CoT, GPT-4 generates an explanation of the error message with a prompt template of ``What part of the PDDL problem do you think is causing this error?.'' 
%GPT-4 then generates the refined problem with the explanation. 
The generated explanation is then added to the input prompt, and GPT-4 generates the refined problem based on it. CR with CoT can be repeated as often as necessary until the planner returns error messages. In the rest of this paper, ViLaIn generates the PDs using CR with CoT unless otherwise specified. Note that ViLaIn performs CR with CoT only if the planner returns an error message.

%%%%% Dataset and Evaluation Metrics
\begin{figure}[t]
    \centering
    \includegraphics[width=0.875\columnwidth]{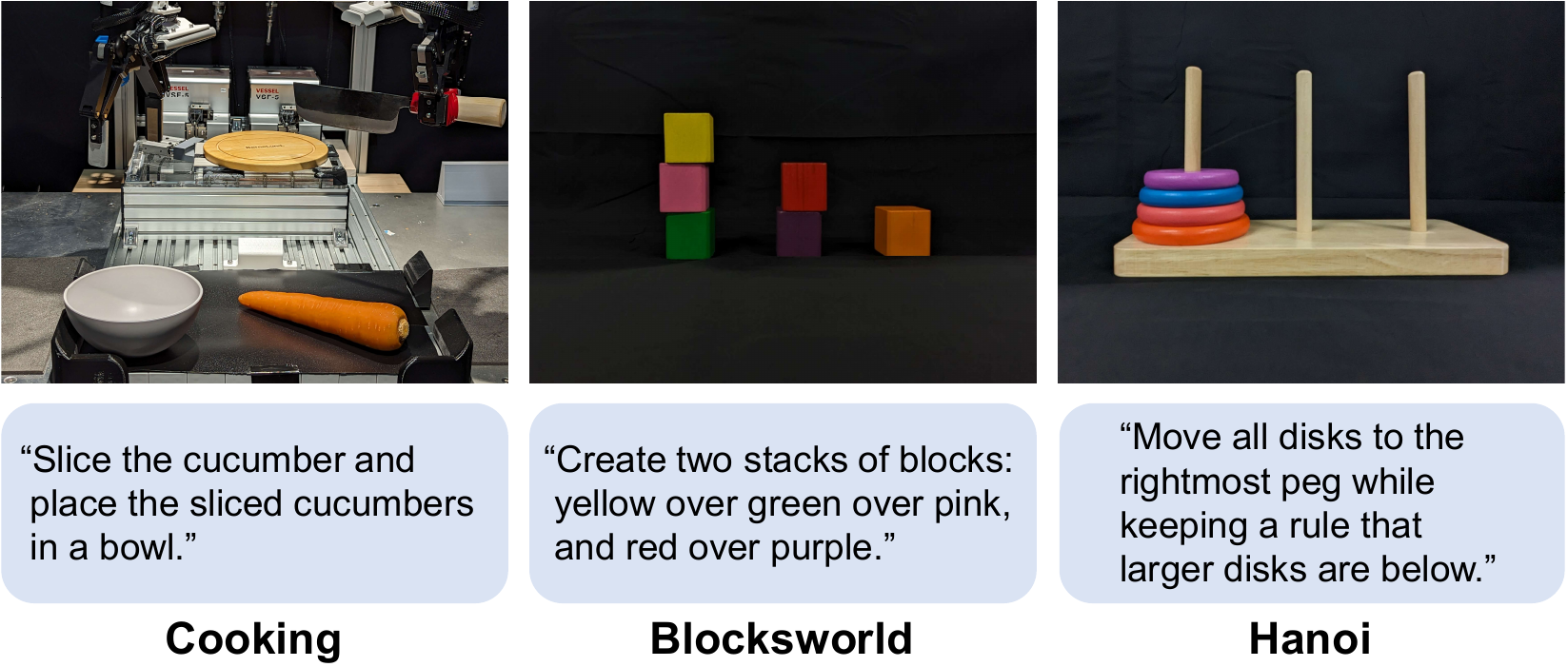}
    \caption{Examples of scene observations and linguistic instructions.}
    \label{fig:examples}
\end{figure}

\section{Dataset}
The evaluation of ViLaIn requires a dataset with linguistic instructions, scene observations, and PDDL domains and problems. However, to our knowledge, no such dataset has been proposed. To this end, we created the ProDG dataset. The ProDG dataset consists of three domains: cooking, the blocks world (Blocksworld), and the tower of Hanoi (Hanoi). 

Cooking is a simplified task of making a salad. Planning is simpler than the other two domains because it only considers slicing vegetables and placing them in the bowl. Cooking actions are supposed to be performed by two robot arms installed on both sides of the environment. The left and right robot arms are named \texttt{a\textunderscore bot} and \texttt{b\textunderscore bot}, respectively, in $O$. This domain handles a greater variety of objects than the other domains. $G$ represents the vegetable state and location. 

Blocksworld is a classical planning domain~\cite{gupta1992complexity}. Fewer types of objects than the cooking appear, but a longer horizon planning is required. Seven colored blocks without duplicates are used for each problem. A robot arm does not hold anything at first. $G$ specifies the relationships of the blocks.

Hanoi is a classical planning domain~\cite{alford2009translating}. Similarly to Blocksworld, a longer horizon planning with fewer types of objects than the cooking domain is required. Ten disks with six colors and three pegs are used. Disks of the same color are named by the number in order of increasing width (e.g., \texttt{blue\textunderscore disk1} and \texttt{blue\textunderscore disk2}). The three pegs are named by the number from left to right (e.g., \texttt{peg1}, \texttt{peg2}, and \texttt{peg3}). $I$ and $G$ specify the positions of the disks. Completing this task requires correctly recognizing the disk sizes since $L$ only instructs the rule of the task, ``larger disks are below,'' but mentions no concrete objects.

Each domain has one domain description and ten PDs. \tabref{tab:domain} shows object types, predicates, and actions in the domain descriptions. Each problem has one linguistic instruction and one scene observation. \figref{fig:examples} shows examples of linguistic instructions $L$ and scene observations $S$. For the Hanoi domain, $L$ is identical through all problems. This aims to investigate whether ViLaIn can generate different $G$ based on $O$ and $I$. The descriptions for the cooking domain were created from scratch, while those for the Blocksworld and Hanoi domains were created based on the PDDL files in pddlgym~\cite{silver2020pddlgym}. We confirmed that all the created PDs are syntactically correct and have solutions using Fast Downward~\cite{helmert2006fast} and VAL, a plan validation software.\footnote{\url{https://github.com/KCL-Planning/VAL}}

\subsection{Evaluation Metrics}\label{subsec:evaluation-metrics}
In PD generation, previously proposed metrics roughly calculate the planning success rate or are domain-specific ones~\cite{liu2023llm,xie2023translating}. It would be ideal to have metrics that evaluate PDs from multiple perspectives regardless of domain. To this end, we introduce a new suite of metrics: \Rsyntax and \Rplan for logical correctness and \Rpart and \Rall for semantic correctness. We describe these metrics below.

\paragraph*{\textbf{\Rsyntax}} PDs must be syntactically correct. \Rsyntax calculates the ratio of such PDs. A PD is considered to be syntactically correct if VAL returns no warnings and exit codes for a pair of the domain and the generated PD.

\paragraph*{\textbf{\Rplan}} Even if the PDs are syntactically correct, they might not have valid plans due to incorrect objects in $O$ and incorrect or contradictory propositions in $I$ and $G$. \Rplan calculates the ratio of the PDs having valid plans. The plans are obtained using Fast Downward~\cite{helmert2006fast}. A plan is considered to be valid if VAL returns no error messages.

\paragraph*{\textbf{\Rpart} and \textbf{\Rall}} The above two metrics ignore whether the PDs are written about our intended tasks. For example, the PD might be about an unintended task while it is syntactically correct and has a valid plan. \Rpart evaluates how close the generated problems are to the ground truth ones. \Rpart calculates the recall of the problem parts between the ground truth and generated ones. \Rpart is independently computed for $O$, $I$, and $G$. The recall of object labels is calculated for $O$, while the recall of propositions is computed for $I$ and $G$. Unlike \Rpart, \Rall calculates the ratio of problems containing all the ground truth object labels and propositions. Thus, \Rall can be viewed as a harder metric than \Rpart.

%%%%% Experiments
\section{Experiments}\label{sec:results}
We conduct experiments to investigate how accurately ViLaIn can generate PDs on the ProDG dataset. This section first describes the generation settings of ViLaIn and then discusses experimental results.

\subsection{Generation Settings of ViLaIn}\label{subsec:generation-settings}
GPT-4 used few-shot prompting with three input/output examples in the same domain as the current task. ViLaIn can refine erroneous PDs by CR $n$ times. PDs with corrected grammatical errors can still have semantic errors, causing no valid solutions. In such cases, CR should be performed at least twice. Thus, we set $n$ to two. For evaluation, we generated ten PDs per problem by varying the example combinations. The resulting 100 problems per domain are used to evaluate ViLaIn.

\subsection{Evaluation of Generation Results by ViLaIn}\label{subsec:main-results}
\begin{table}[t]
    \centering
    \caption{Performance on the ProDG dataset} 
    \begin{tabular}{l|c|c|c|c|c|c} \hline
        \multicolumn{1}{c|}{\multirow{2}{*}{Domain}} & \multicolumn{1}{c|}{\multirow{2}{*}{\Rsyntax}} & \multicolumn{1}{c|}{\multirow{2}{*}{\Rplan}} & \multicolumn{3}{c|}{\Rpart} & \multicolumn{1}{c}{\multirow{2}{*}{\Rall}} \\\cline{4-6}
        & & & \multicolumn{1}{c|}{$O$} & \multicolumn{1}{c|}{$I$} & \multicolumn{1}{c|}{$G$} & \\\hline
        Cooking & 0.99 & 0.99 & 1.00 & 0.93 & 0.93 & 0.71 \\
        Blocksworld & 0.99 & 0.94 & 0.98 & 0.79 & 0.89 & 0.36 \\
        Hanoi & 1.00 & 0.58 & 0.89 & 0.46 & 0.33 & 0.12 \\\hline
    \end{tabular}
    \label{tab:main-results}
\end{table}

\tabref{tab:main-results} shows the results. The \Rsyntax scores are more than 99\% in all the three domains. This means that ViLaIn can generate syntactically correct PDs for these domains utilizing the three input/output examples. The \Rplan scores indicate that 94\% or more PDs have valid plans in the cooking and Blocksworld domains. However, in the Hanoi domain, the \Rplan score is only 58\% due to its challenging setting. We found from the outputs that ViLaIn tends to omit some propositions in this domain, making the PDs invalid. 

For \Rpart, the scores on $I$ and $G$ are smaller than those on $O$. This implies that generating $I$ and $G$ is more challenging than $O$. We found that mistakenly detected objects cause this. Predicates such as \texttt{on} or \texttt{at} take two objects as arguments. Propositions created with the predicates and mistakenly detected objects affect other propositions. For example, \texttt{(on red\textunderscore block blue\textunderscore block)} can be \texttt{(on red\textunderscore block green\textunderscore block) (on green\textunderscore block blue\textunderscore block)} with a mistakenly detected \texttt{green\textunderscore block}, making them all incorrect propositions. We consider that generating these incorrect propositions causes such results.

Finally, the \Rall score is 71\% in the cooking domain, 36\% in the Blocksworld domain, and 12\% in the Hanoi domain. The scores in the cooking and Hanoi domains make sense considering the \Rplan and \Rpart scores. However, the score is unexpectedly low in the Blocksworld domain. We found that PDs in the Blocksworld domain tend to contain a few incorrect propositions of block relationships. In some cases, the block positioning is mistakenly reversed (e.g., \texttt{(on blue\textunderscore block red\textunderscore block) (on red\textunderscore block green\textunderscore block)} is reversed to \texttt{(on green\textunderscore block red\textunderscore block) (on red\textunderscore block blue\textunderscore block)}). We consider that these lead to the low \Rall score in this domain. 

\begin{figure}[t]
    \centering
    \includegraphics[width=0.9\columnwidth]{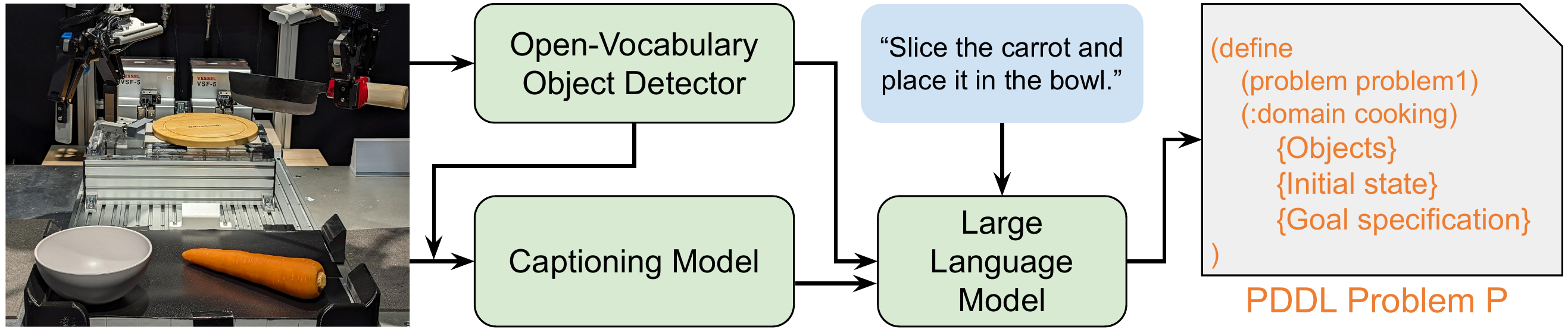}
    \caption{\Vilainwhole generates the whole problem description at once.}
    \label{fig:vilain-whole}
\end{figure}

\begin{table}[t]
    \centering
    \caption{Generating the whole problem descriptions at once}
    \begin{tabular}{l|l|l|l} \hline
        \multicolumn{1}{c|}{Domain} & \multicolumn{1}{c|}{\Rsyntax} & \multicolumn{1}{c|}{\Rplan} & \multicolumn{1}{c}{\Rall} \\\hline
        Cooking & 1.00 (+0.01) & 1.00 (+0.01) & 0.54 (-0.17) \\ 
        Blocksworld & 0.99 (+0.00) & 0.99 (+0.05) & 0.13 (-0.23) \\ 
        Hanoi & 1.00 (+0.00) & 0.94 (+0.36) & 0.21 (+0.09) \\\hline
    \end{tabular}
    \label{tab:whole-results}
\end{table}

\subsection{Generating the Whole Problem at Once}
ViLaIn generates the parts of PDs using different modules. If a single module can generate the whole problem at once, it greatly simplifies the system. Here, we consider a variant of ViLaIn generating the whole PD at once, as illustrated in \figref{fig:vilain-whole}. We refer to this model as \Vilainwhole. The generation is performed with few-shot prompting as the original model.

\tabref{tab:whole-results} shows the results with \Rsyntax, \Rplan, and \Rall. Values inside parenthesis indicate gains from ViLaIn. In the cooking and Blocksworld domains, \Vilainwhole slightly improves \Rplan but worsens \Rall. This means that using three modules is more effective for these domains. In the Hanoi domain, \Vilainwhole outperforms ViLaIn in both \Rplan and \Rall. When considered with \secref{subsec:main-results}, this means that \Vilainwhole generates more correct propositions than ViLaIn. Generating the whole PDs makes the distance between tokens of $O$ and $I$ or $G$ closer. We consider that this might work effectively and result in these improvements.

\begin{table}[t]
    \centering
    \caption{Performance without CR and CoT}
    \begin{tabular}{c|c|c|c|c} \hline
        \multicolumn{2}{c|}{CR configurations} & \multirow{2}{*}{\Rsyntax} & \multirow{2}{*}{\Rplan} & \multirow{2}{*}{\Rall} \\\cline{1-2}
        CR ($n$ times) & CoT & & &  \\\hline
        2 & \checkmark & 0.99 & 0.99 & 0.71 \\
        1 & \checkmark & 0.99 & 0.94 & 0.68 \\
        1 & & 0.97 & 0.85 & 0.59 \\
        0 & & 0.60 & 0.18 & 0.09 \\\hline
    \end{tabular}
    \label{tab:ablation}
\end{table}

\subsection{Generating PDs without CR and CoT}
ViLaIn uses corrective re-prompting (CR) and chain-of-thought (CoT) prompting. The CR is performed twice at most as described in \secref{subsec:generation-settings}. Since all the PDs so far are generated using CR with CoT, the impact of CR on performance is still unknown. Here, we investigate performance without CR and CoT, considering the following configurations: (i) CR with CoT ($n$ = 1 in \secref{subsec:generation-settings}), (ii) CR without CoT ($n$ = 1), and (iii) without CR ($n$ = 0).

\tabref{tab:ablation} shows the results in the cooking domain. The first line is the same result in \tabref{tab:main-results}. First, performing CR with CoT only once (the first line) slightly drops \Rplan and \Rall, meaning that repeating CR is effective. Next, removing CoT (the third line) worsens all the scores. This demonstrates that the introduced intermediate reasoning step by CoT has a large impact on performance. Finally, removing CR (the fourth line) degrades the scores significantly. This model tends to suffer from \textit{hallucinations}~\cite{maynez2020faithfulness}\footnote{Also referred to as \textit{confabulations}. Generating factually incorrect texts by LLMs is a common problem in natural language processing.}, such as propositions with undefined objects (e.g., \texttt{(at cucumber counter)} in $I$ while the cucumber is not defined in $O$). We found that CR effectively refines these incorrect propositions and makes the PDs consistent.

%%%%% Conclusion
\section{CONCLUSION}\label{sec:conclusion}
This paper has tackled multimodal planning problem specification, a new task for connecting language-guided planning and symbolic planner. We have proposed Vision-language interpreter (ViLaIn) that generates problem description (PD)s from linguistic instructions and scene observations. A novel dataset called the problem description generation (ProDG) dataset has proposed with new metrics to evaluate ViLaIn. The experimental results show that ViLaIn can generate syntactically correct PDs and more than half of the PDs have valid plans. Interesting future directions include (i) constructing a robotic system with ViLaIn that executes linguistic instructions, (ii) refining PDs via errors from real robots, and (iii) reducing human effort for new tasks.

\section*{ACKNOWLEDGMENT}
We would like to thank Hirotaka Kameko for his helpful comments. This work was supported by JSPS KAKENHI Grant Number 20H04210 and 21H04910 and JST Moonshot R\&D Grant Number JPMJMS2236.

%%%%%%%%%%%%%%%%%%%%%%%%%%%%%%%%%%%%%%%%%%%%%%%%%%%%%%%%%%%%%%%%%%%%%%%%%%%%%%%%

{
    \bibliographystyle{IEEEtran}
    \bibliography{main}
}

\end{document}